\documentclass[conference]{IEEEtran}
\IEEEoverridecommandlockouts

\usepackage{cite}
\usepackage{amsmath,amssymb}
\usepackage{graphicx}
\usepackage[T1]{fontenc}
\usepackage[utf8]{inputenc}
\usepackage{textcomp}
\usepackage{xcolor}

\usepackage[hidelinks,breaklinks]{hyperref}

\begin{document}

\title{Pattern Intelligence: Unsupervised, LLM-Interpreted\\
Discovery of Systematic Underpayment and Denial Patterns\\
in Healthcare Revenue Cycle Management}

\author{%
\IEEEauthorblockN{Harshil Lodhiya, Alex McManus}
\IEEEauthorblockA{\textit{SlicedHealth}\\
Woodstock, Georgia, United States\\
hlodhiya@slicedhealth.com; amcmanus@slicedhealth.com}%
\thanks{Accepted for publication in the 2027 IEEE Gujarat Section Conference
(GSCON 2027). \copyright{}~2027 IEEE. Personal use of this material is
permitted. Permission from IEEE must be obtained for all other uses, in any
current or future media, including reprinting/republishing this material for
advertising or promotional purposes, creating new collective works, for resale
or redistribution to servers or lists, or reuse of any copyrighted component of
this work in other works.}%
}

\maketitle

\begin{abstract}
Healthcare providers lose substantial revenue to systematic payer
underpayments and claim denials, yet identifying the recurring, correctable
patterns behind them remains largely manual and does not scale to the hundreds
of thousands of claims a hospital adjudicates. We present Pattern
Intelligence, a deployed system that discovers such patterns without
supervision by clustering claims on their adjudication behaviour rather than
their identity. Each claim is embedded in a 130-dimensional feature vector
combining per-line expected-versus-actual payment ratios, claim adjustment
reason (CARC) codes, denial and payment signals, and service composition.
Payer and contract identity are deliberately excluded from the distance
metric --- we show they dominate the geometry and collapse clusters into
trivial per-payer groups --- and are reintroduced only as a post-hoc
segmentation (by contract and visit type) that keeps every surfaced pattern
coherent and actionable. We cluster with HDBSCAN, recover additional structure
from the noise set via a second-pass sub-clustering, and gate low-confidence
memberships so that a pattern's claims genuinely represent it. A large
language model translates each cluster centroid into an analyst-ready
description --- dominant procedures, adjustment reasons, and a plain-language
failure mode. We report a recurring empirical finding that a substantial
fraction of high-dollar denials --- between roughly one-third and one-half,
depending on tenant and segmentation --- forms an irreducible, idiosyncratic
long tail that no density-clustering configuration or representation we tried
groups, and argue this reflects a property of the domain rather than a
modeling deficiency. We evaluate pattern validity through cluster-coherence
analysis (per-pattern homogeneity before and after segmentation) and quantify
impact with a recoverable-opportunity (\$) metric, reported on two production
tenants of contrasting scale ($\approx$10K and $\approx$157K claims). We
further describe a multi-tenant deployment, currently across six hospital
systems (deployment expanding) at ${\sim}$10K--157K claims each. Rather than a
new algorithm, the contribution is a problem formulation, a
representation-design insight, and empirical findings from a production system
on real claims data.
\end{abstract}


\section{Introduction}

Revenue cycle management (RCM) is the set of administrative and financial
processes by which healthcare providers are reimbursed for care. A material
fraction of billed charges is never fully collected: payers deny claims, apply
contractual adjustments, or reimburse below the expected contracted rate. Many
of these shortfalls are not random --- they recur systematically for a given
payer, service, contract term, or coding pattern, and are therefore
correctable. Identifying which shortfalls are systematic (and thus worth an
appeal, a contract renegotiation, or a coding fix) is today performed by
analysts reviewing claims largely one at a time, a process that does not scale
to the volume a modern hospital adjudicates.

Machine learning applied to healthcare has concentrated on clinical
prediction; the operational and financial layer of RCM --- denials,
underpayments, and their systematic causes --- is comparatively under-studied.
The task is a poor fit for supervised learning: there is no reliable label for
``this is a systematic, correctable pattern,'' the space of failure modes is
open-ended and drifts as payers change policy, and any labeling scheme would
encode the very analyst intuition we wish to scale. We therefore frame the
problem as unsupervised discovery: surface coherent groups of claims that fail
(or under-pay) for the same reason, and present each group to an analyst in
actionable terms.

This paper describes Pattern Intelligence, a system deployed in production
across multiple hospital tenants. Our contributions are:

\begin{itemize}
\item A problem formulation for unsupervised discovery of systematic
underpayment and denial patterns in RCM claims, distinct from clinical
prediction and from supervised denial-propensity models.

\item A representation-design insight: clustering on adjudication
\textit{behaviour} while deliberately excluding payer and contract identity
from the distance metric, because those features dominate the geometry and
shatter clusters trivially; identity is reintroduced as post-hoc segmentation.

\item A coherence mechanism --- contract-and-visit segmentation plus
confidence gating and a homogeneity criterion --- that keeps every surfaced
pattern one an analyst can act on without dismissing it as mixed data.

\item An LLM interpretation layer that converts cluster centroids into
analyst-ready, coded (CARC-referenced) descriptions and failure modes.

\item Empirical findings from a real deployment, including an honest negative
result: an irreducible long tail of high-dollar, idiosyncratic denials that
resists density clustering under every configuration we tried.
\end{itemize}

\section{Related Work}

\textit{Denials and RCM analytics.} Prior work on claim denials is dominated
by supervised denial-propensity or payment-prediction models (e.g.,
\cite{kumar2010}, \cite{kim2020}, \cite{johnson2023}). These require labeled
outcomes and predict \textit{whether} a claim will be denied rather than
surfacing the \textit{systematic, correctable structure} across many claims.
Our aim is complementary and unsupervised.

\textit{Density-based clustering.} We build on DBSCAN \cite{ester1996} and its
hierarchical density-based extension HDBSCAN \cite{campello2013},
\cite{mcinnes2017}, which does not require a preset number of clusters and
yields a principled noise label --- both essential when the number and
prevalence of failure modes are unknown a priori. Cluster-validity for density
methods (e.g., DBCV \cite{moulavi2014}) informs our parameter selection.

\textit{Representation for tabular/administrative data and interpretation.}
Representation and feature engineering for tabular and administrative data is
an active area \cite{miotto2016}, \cite{gorishniy2021}. Recent work uses large
language models to label or summarize clusters and topics \cite{wang2023},
\cite{pham2024}; adaptive routing strategies for selecting among language
models under non-stationary workloads have also been studied
\cite{lodhiya2026}. We apply this specifically to translate claim-cluster
centroids into coded, operationally actionable descriptions.

\section{Problem Formulation}

Let a claim be adjudicated at the service-line level, producing for each line
an expected (contracted) payment, an actual payment, and zero or more
adjustment reason (CARC) codes with group codes and amounts. A claim carries a
payer, a contract, a visit type (e.g., professional, outpatient), and derived
financial quantities (charged, expected, actual, variance). We seek a
partition of a tenant's claims into groups such that (i) claims in a group
share an adjudication \textit{behavior} --- the same manner of underpayment or
denial --- and (ii) each surfaced group is homogeneous enough in payer,
contract, and visit that an analyst can act on it as a unit. Groups need not
cover all claims: claims with no shared, recurring behavior should be labeled
noise rather than forced into a group.

\section{Methods}

\subsection{Behavior-based representation}

Each claim is embedded in a 130-dimensional vector composed of blocks: (a)
financial-behavior ratios (expected-vs-actual, variance-to-charge,
underpayment fraction), gated by a significant-variance threshold so that
net-even redistributions do not read as underpayments; (b) a multi-hot
encoding of the most frequent CARC adjustment codes; (c) per-line signals
derived from the estimated payment details (procedure, revenue code, DRG,
diagnosis presence, line-level behavior); and (d) service-group composition.
Continuous blocks are load-bearing: purely categorical variants degraded
cluster quality in our experiments.

Crucially, payer and contract identity are excluded from the vector used for
the distance metric. When included, they dominate Euclidean distance and cause
HDBSCAN to partition trivially by payer, destroying the behavioral signal we
want. We instead retain identity as metadata and reintroduce it post hoc
(Section~\ref{subsec:coherence}). A frozen encoder --- vocabularies of the
top-$k$ codes and global financial min/max --- guarantees that every claim in
a tenant is embedded in the same feature space, a prerequisite for comparable
distances; the encoder is fit once over the full corpus and persisted.

\begin{figure}[!t]
\centering
\includegraphics[width=\columnwidth]{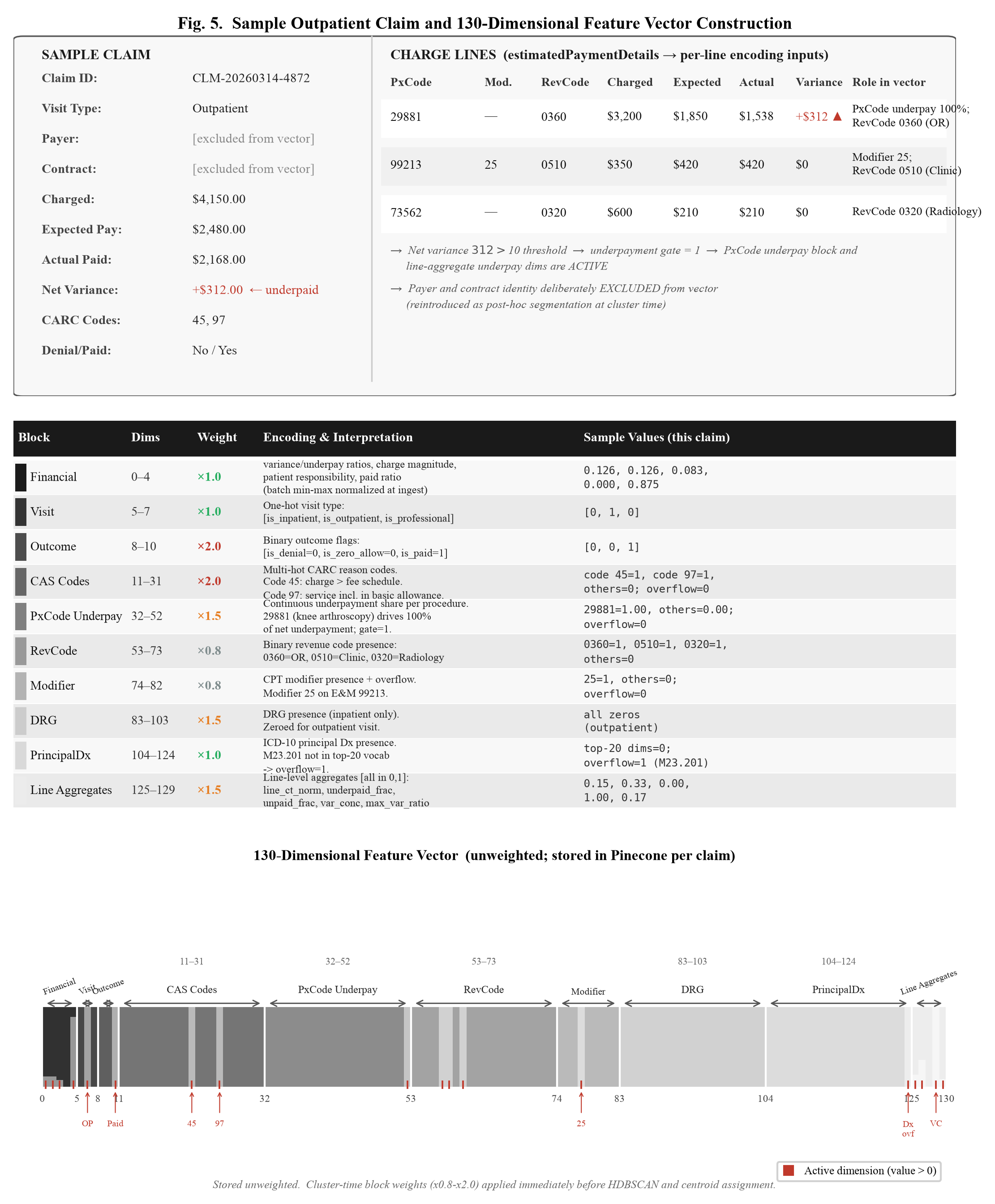}
\caption{Sample outpatient claim and 130-dimensional feature vector
construction. The claim card (top) shows the claim-level and per-line inputs;
the table (middle) gives the block layout, cluster-time weights, and the
encoded values for this claim; the strip (bottom) shows the resulting vector
with active dimensions marked.}
\label{fig:vector}
\end{figure}

\subsection{Density clustering}

We cluster with HDBSCAN, which infers the number of clusters from density
structure and assigns a noise label to points with no dense neighborhood.
Parameters (minimum cluster size, minimum samples, selection method) are
derived from tenant size, with a cap so that a value sensible for a large
tenant does not force all-noise on a smaller one; we optionally auto-tune the
minimum cluster size using a density-based validity score with a NaN-safe
fallback.

\subsection{Coherence: segmentation, noise recovery, and confidence gating}
\label{subsec:coherence}

A behavioral cluster may legitimately span payers and visit types, which is
undesirable when the surfaced pattern must be acted on per contract. We
therefore segment each cluster by (contract, visit type): a claim's pattern is
its behavioral cluster intersected with its contract and visit. This
guarantees that the payer and visit shown on a surfaced pattern describe every
claim in it, eliminating the failure mode in which, e.g., an out-of-family
outpatient claim appears under a professional pattern labeled for a different
payer.

We run a second-pass sub-clustering over the noise set to recover structure
that the global pass missed, applying the same segmentation so that recovered
patterns are equally coherent. Finally, we gate low-confidence memberships:
claims whose density-membership probability falls below a threshold are
returned to noise rather than assigned, so a pattern's reported membership
equals its analyst-visible membership and marginal claims do not contaminate a
pattern's characterization.

\subsection{LLM interpretation}

Cluster geometry is opaque to an analyst. For each pattern we decode the
centroid into its dominant driver codes and line-level behavior, translate the
associated CARC codes into human-readable reasons, classify a failure mode
(e.g., denial vs.\ contractual zero-allowable vs.\ underpayment), and prompt a
large language model to produce a short, specific pattern name and an
analyst-ready summary grounded in these decoded signals. Names are
disambiguated so distinct patterns do not collide. The result turns a dense
centroid into concrete review guidance.

\section{System and Deployment}

Pattern Intelligence runs multi-tenant. A historical ingest streams a tenant's
claims from the source system, fits the encoder, embeds and stores vectors,
then clusters and writes back each claim's pattern assignment and each
pattern's description. An incremental daily job embeds new claims and assigns
them to the nearest existing pattern; periodic housekeeping and reconciliation
keep the vector store consistent with the source of truth. The pipeline is
checkpointed and resumable, and coverage is summarized to analysts via an
anomalous-claim percentage and a recoverable-opportunity dollar metric. At the
time of writing the system is deployed across six hospital tenants, with
deployment expanding, processing approximately 10K--157K claims each.

\section{Evaluation}

Evaluating unsupervised discovery on proprietary claims data is the central
difficulty: there is no gold-standard partition. We address it two ways: an
internal analysis of cluster coherence and the noise floor, and an
operational, dollar-weighted measure of impact. All figures below are measured
on two production tenants of contrasting scale --- Tenant~A (157,184 claims)
and Tenant~B (10,318 claims).

\subsection{Coherence and the irreducible long tail}
\label{subsec:coherence-eval}

We measure per-pattern homogeneity as the claim-weighted fraction of clustered
claims that match their pattern's dominant (contract, visit type) pair. Before
segmentation, the behavioral clusters are already largely coherent --- 92.7\%
on Tenant~B and 66.3\% on the larger, more heterogeneous Tenant~A ---
confirming that a purely behavioral cluster can nonetheless span contracts and
visit types. After (contract, visit) segmentation, homogeneity is 100\% by
construction on both tenants: every surfaced pattern is, by definition,
actionable as a single contract-and-visit unit. This coherence is not free.
Segmentation returns contract-and-visit fragments below the minimum segment
size to noise, so the unclustered fraction rises (Tenant~A
22.9\%$\rightarrow$45.4\%, Tenant~B 30.4\%$\rightarrow$32.5\%) and the share of
recoverable dollars captured in coherent patterns falls
(Section~\ref{subsec:impact}). The design therefore trades coverage for
guaranteed actionability, and the cost is larger for the large, heterogeneous
tenant.

\begin{figure}[!t]
\centering
\includegraphics[width=\columnwidth]{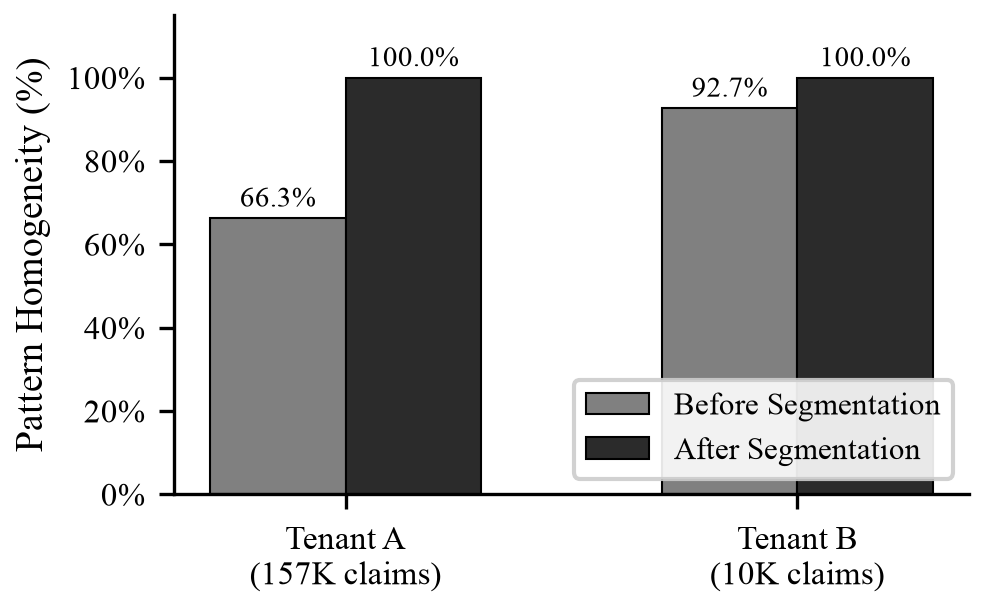}
\caption{Per-pattern homogeneity before and after segmentation. Behavioral
clusters are already largely coherent; segmentation brings homogeneity to
100\% by construction.}
\label{fig:homogeneity}
\end{figure}

A substantial fraction of claims remains unclustered under every configuration
we evaluated --- 22.9\%--45.4\% depending on tenant and segmentation, and,
viewed on anomalous (denial / zero-allowable / high-variance) claims
specifically, 34.8\%--54.5\% of them fall outside any pattern
post-segmentation. This residue is concentrated in high-dollar, idiosyncratic
denials. Two sweeps indicate it is a property of the data rather than a tuning
artifact. First, varying the HDBSCAN minimum cluster size does not dislodge
it: on Tenant~B the noise floor stays in a narrow 26.2\%--32.8\% band across
minimum-cluster-size candidates from 20 to 62, and on Tenant~A it is
essentially flat (22.8\%--22.9\%) across candidates 100 and 200. Second,
representation matters in the expected direction but does not remove the
floor: dropping the continuous financial and per-line behavior blocks to a
categorical-only vector on Tenant~B raised noise
(30.4\%$\rightarrow$32.8\%) and coarsened the solution
(92$\rightarrow$73 clusters), confirming that the continuous behavior blocks
are load-bearing and that no representation we tried clusters the tail. We
therefore argue the actionable objective is dollar-weighted triage rather than
exhaustive coverage.

\begin{figure}[!t]
\centering
\includegraphics[width=\columnwidth]{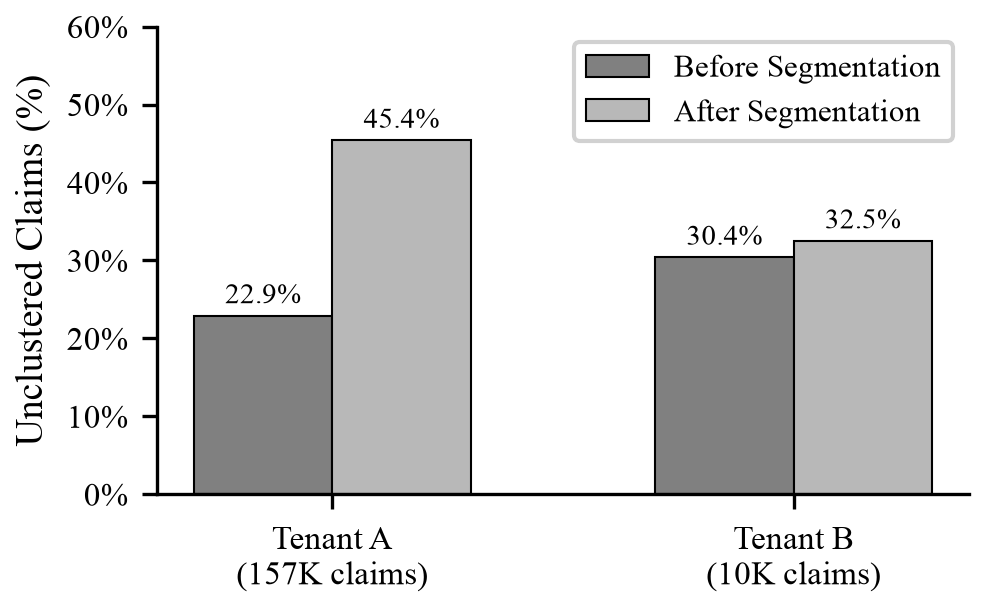}
\caption{Unclustered claim fraction before and after (contract, visit)
segmentation. Segmentation trades coverage for guaranteed per-contract
actionability.}
\label{fig:unclustered}
\end{figure}

\begin{figure}[!t]
\centering
\includegraphics[width=\columnwidth]{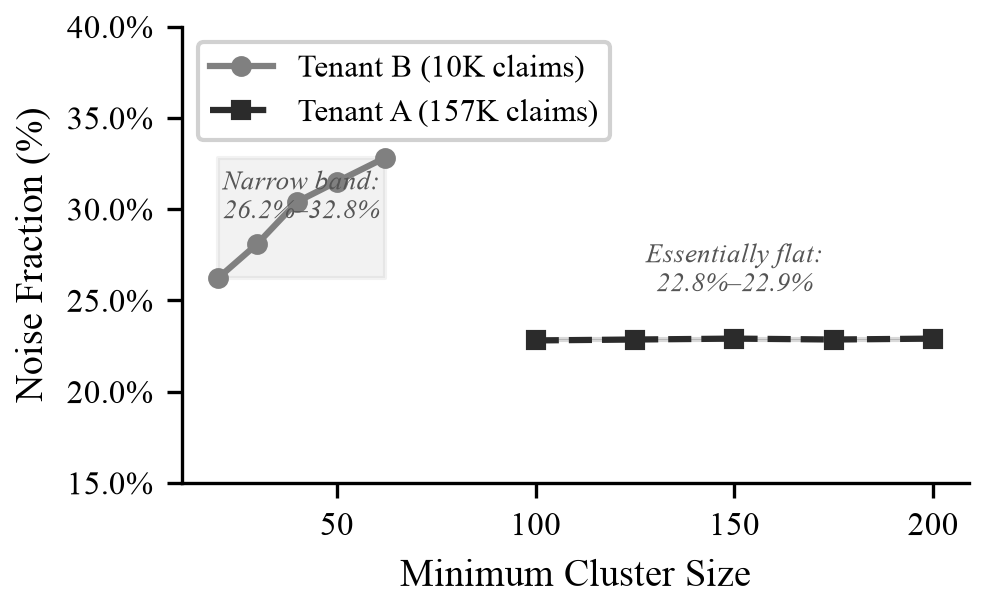}
\caption{Noise fraction as a function of HDBSCAN minimum cluster size. The
floor is stable across a wide parameter range, indicating an irreducible
property of the data.}
\label{fig:noisefloor}
\end{figure}

\subsection{Operational impact}
\label{subsec:impact}

We quantify impact as recoverable opportunity: the sum of positive net
variance on anomalous claims (denials, zero-allowables, and claims whose net
variance exceeds a significance threshold), and the fraction of those dollars
that lands in a coherent, analyst-visible pattern. Tenant~A carries \$9.63M in
recoverable opportunity across 55,648 anomalous claims; Tenant~B carries
\$1.20M across 5,912. Behavioral clustering alone captures 25.3\% of that
opportunity on Tenant~A and 41.7\% on Tenant~B in analyst-reviewable groups.
After (contract, visit) segmentation --- the form in which patterns are
actually surfaced --- capture is 13.8\% (\$1.33M) on Tenant~A and 39.0\%
(\$469K) on Tenant~B. The gap between the two tenants, and the drop induced by
segmentation, both reflect the same underlying fact established in
Section~\ref{subsec:coherence-eval}: the large tenant's recoverable dollars
are more dispersed across one-off, high-dollar denials, so more of them fall
in the irreducible tail. We report opportunity surfaced rather than dollars
recovered; downstream appeal and recovery outcomes are not tracked within the
system and are left to future work. Even so, capturing on the order of \$1.3M
(Tenant~A) and \$0.47M (Tenant~B) of concentrated, contract-attributable
opportunity per tenant is the operationally relevant result, and it is
consistent with a triage objective that prioritizes dollar-dense patterns over
exhaustive coverage.

\begin{figure}[!t]
\centering
\includegraphics[width=\columnwidth]{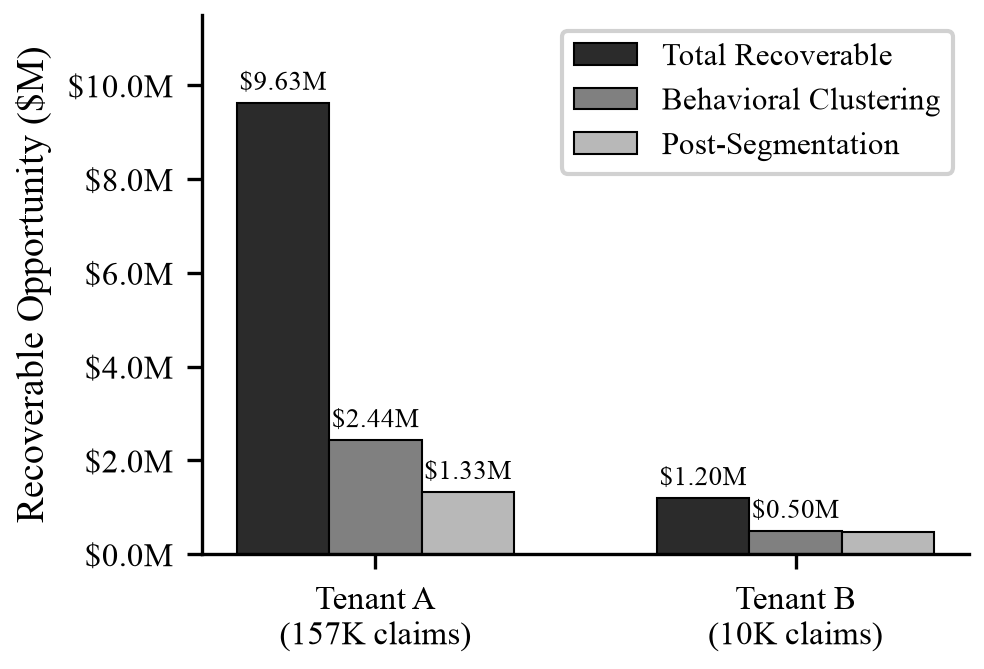}
\caption{Recoverable opportunity by pipeline stage. Total recoverable dollars,
the fraction captured by behavioral clustering alone, and the fraction
retained after (contract, visit) segmentation.}
\label{fig:opportunity}
\end{figure}

\section{Discussion}

The central design tension is that the signal we want (shared adjudication
behavior) and the axis analysts act on (payer/contract/visit) are different:
optimizing distance on the latter destroys the former. Separating them ---
behavior in the metric, identity in segmentation --- resolves the tension and
is, we believe, transferable to other administrative-data settings where a
dominant categorical field would otherwise capture the clustering. The
confidence gate and homogeneity criterion encode a product requirement often
left implicit in clustering papers: a surfaced group is only useful if a user
can trust that its members belong together.

\section{Limitations and Ethical Considerations}

\begin{itemize}
\item \textit{Evaluation without ground truth.} Absent a gold partition, our
validity evidence is indirect (cluster-coherence analysis and dollar-weighted
opportunity outcomes) and is drawn from two tenants; we do not claim to
recover a `true' clustering, and controlled fault-injection recovery and
blinded analyst adjudication remain future work.

\item \textit{Data.} Results derive from proprietary claims data across a
limited set of tenants and one vendor pipeline; generalization to other
providers and payers is untested. The data cannot be released.

\item \textit{Privacy.} All analysis is on de-identified, aggregate quantities
derived from claims data; no protected health information is accessed or
reported, and no individual claim is disclosed. This study is a retrospective
analysis of de-identified, aggregated operational claims data and does not
constitute human-subjects research; institutional review board review was
therefore not required.

\item \textit{Methods.} The components are established (HDBSCAN, LLM
prompting); the contribution is applied --- formulation, representation
design, and deployment findings --- not a new algorithm.
\end{itemize}

\section{Conclusion}

We described a deployed system that discovers systematic underpayment and
denial patterns in healthcare claims without supervision by clustering on
adjudication behavior, keeping payer and contract identity out of the distance
metric and reintroducing them as segmentation, and interpreting clusters with
an LLM. We reported an honest limit --- an irreducible idiosyncratic long tail
--- and argued for dollar-weighted triage over exhaustive coverage. We hope
the representation-design insight and the deployment findings are useful to
others applying unsupervised learning to administrative healthcare data.

\section*{Competing Interests}

Harshil Lodhiya and Alex McManus are employed by SlicedHealth, the company
that develops the Pattern Intelligence system described in this manuscript.
The authors hold stock options in SlicedHealth. The authors have no other
competing interests to declare.

\section*{Funding}

This work was funded by SlicedHealth, Woodstock, Georgia, United States, which
provided support in the form of a salary for authors H.L. and A.M., as well as
the claims data and computational resources used in the study. As the authors
are employees of SlicedHealth, the funder was involved in the study design,
analysis, and preparation of the manuscript.

\section*{Data Availability}

The claims data underlying this study are proprietary and contain protected
health information; they cannot be shared publicly. The aggregate and derived
statistics reported in this manuscript are available from the corresponding
authors on reasonable request, subject to a data-use agreement. The Pattern
Intelligence software described here is proprietary to SlicedHealth and is not
publicly available.

\end{document}